\documentclass[10pt,twocolumn,letterpaper]{article}

\usepackage{iccv}
\usepackage{times}
\usepackage{epsfig}
\usepackage{graphicx}
\usepackage{amsmath}
\usepackage{amssymb}
\usepackage{color} 
\usepackage[ruled]{algorithm2e}
\usepackage{multirow} 

\usepackage[pagebackref=true,breaklinks=true,letterpaper=true,colorlinks,bookmarks=false]{hyperref}

\iccvfinalcopy 

\def\iccvPaperID{6520} 

\ificcvfinal\fi

\begin{document}

\title{Densely Guided Knowledge Distillation using Multiple Teacher Assistants}

\author{Wonchul Son, Jaemin Na, Junyong Choi, Wonjun Hwang\\
Ajou University, Republic of Korea\\
{\tt\small \{dnjscjf92, osial46, chldusxkr, wjhwang\}@ajou.ac.kr}
}

\maketitle
\ificcvfinal\thispagestyle{empty}\fi

\begin{abstract}
    With the success of deep neural networks, knowledge distillation which guides the learning of a small student network from a large teacher network is being actively studied for model compression and transfer learning. 
    However, few studies have been performed to resolve the poor learning issue of the student network when the student and teacher model sizes significantly differ.
    In this paper, we propose a densely guided knowledge distillation using multiple teacher assistants that gradually decreases the model size to efficiently bridge the large gap between the teacher and student networks.
    To stimulate more efficient learning of the student network, we guide each teacher assistant to every other smaller teacher assistants iteratively. 
    Specifically, when teaching a smaller teacher assistant at the next step, the existing larger teacher assistants from the previous step are used as well as the teacher network.
    Moreover, we design stochastic teaching where, for each mini-batch, a teacher or teacher assistants are randomly dropped. This acts as a regularizer to improve the efficiency of teaching of the student network.
    Thus, the student can always learn salient distilled knowledge from the multiple sources.
    We verified the effectiveness of the proposed method for a classification task using CIFAR-10, CIFAR-100, and ImageNet. We also achieved significant performance improvements with various backbone architectures such as ResNet, WideResNet, and VGG.
    \footnote{
    Our code is available at https://github.com/wonchulSon/DGKD.}
\end{abstract}

\section{Introduction}
\label{sec:Introduction}

\begin{figure}[t]
\centering
\includegraphics[width=8.3cm]{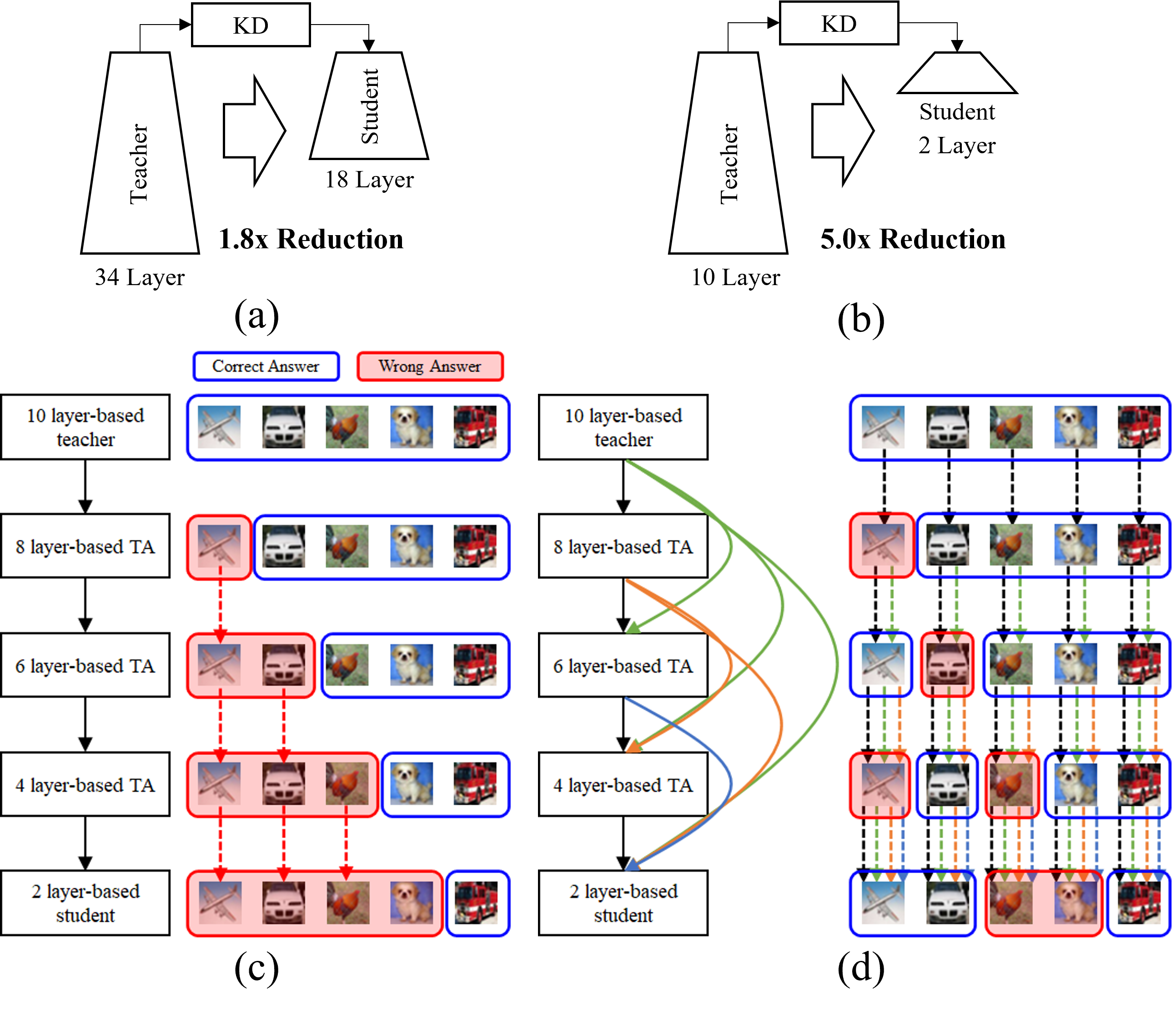}
\caption{\textbf{Problem definition of the large gap between a teacher and a student network}. (a) In general, the difference between layers at KD is approximately 1.8 times, but (b) we are interested in the challenging problem of layer differences of more than 5 times. For solving this problem, TAKD~\cite{TAKD2020} has been proposed. However, (c) TAKD has a fundamental limitation such as the error avalanche problem. 
Assuming that a unique error occurs one by one when a higher-level teacher assistant (TA) teaches a lower-level TA.
The error case continues to increase whenever teaching more TAs. Meanwhile, in (d), the proposed densely guided knowledge distillation can be relatively free from this error avalanche problem because it does not teach TAs at each level alone.}
\label{fig:01}
\vspace{-2mm}
\end{figure}

\noindent While deep learning-based methods \cite{ResNet, DenseNet, MaskRCNN, SegNet}, e.g., convolutional neural networks (CNNs), have achieved very impressive results in terms of accuracy, there have been many trials \cite{Pruning, MobileNet, Jiaxiang2016, HintonKD} to apply them to many applications such as classification, detection, and segmentation. Among these attempts, Knowledge Distillation (KD) \cite{HintonKD, FitNets} transfers the knowledge of a teacher model (e.g., a deeper or wider network) in the form of soft probability (e.g., logits) to improve the accuracy of a less-parameterized student model (e.g., a shallower network) during a learning procedure. Specifically, the soft logits of the teacher network can train the student network more efficiently than the softmax based on the class label of the student network itself. Many studies \cite{HintonKD, FitNets, Zagoruyko2017, FSP, Mutual, Tung2019, DDKD2020} on the KD method have been proposed, most of which focused on effectively guiding a teacher’s soft probability or outputs to a student. Recently, there have been ensemble-based attempts \cite{Defang2020, Linfeng2019, Yuan2020, Malinin2020} to train a student network based on many peers or students without considering the single teacher network, which is a slight lack of consideration for the diversity of ensemble classifiers teaching the student, especially when the gap between a teacher and a student is large like Figure~\ref{fig:01} (a) and (b).

In \cite{JangHyun2019, Xiao2019}, it was shown that a teacher does not always have to be smart for a student to learn effectively. The KD can not succeed when the student's capacity is too low to successfully mimic the teacher’s knowledge. Recently, to overcome this problem, TA-based knowledge distillation (TAKD)~\cite{TAKD2020} using intermediate-sized assistant models was introduced to alleviate the poor learning of a student network when the size gap between a student and a teacher is large. It achieved an effective performance improvement in the case of a large gap in teacher and student sizes. However, further studies are required to determine whether using middle-sized assistant models in series is the most efficient KD method for bridging the gap between a teacher and a student.  
For example, TAKD tends to cause the error avalanche problem, as shown in Figure \ref{fig:01} (c). It sequentially trains the multiple TA models by decreasing the capacity of their own assistant models for efficient student learning. If an error occurred during a specific TA model learning, this TA model will teach the next level assistant models including the same error. From then on, each time a TA is trained, the error snowballs gradually, as illustrated in Figure~\ref{fig:01} (c). This error avalanche problem becomes an obstacle to improving the student model’s performance as the total number of TAs increases. 

In this paper, we propose a novel densely guided knowledge distillation (DGKD) using multiple TAs for efficient learning of the student model despite the large size gap between a teacher and a student model. As shown in Figure~\ref{fig:01} (d), unlike TAKD, when learning a TA whose model size gradually decreases for the target student, the knowledge is not distilled only from the higher-level TA but guided from all previously learned higher-level TAs including the teacher. Thus, a trainee had distilled knowledge by considering the relationship between the multiple trainers (e.g., teacher and TAs) with complementary characteristics. The error avalanche problem could be alleviated successfully. It is largely because the distilled knowledge previously used for the teaching of models disappears in TAKD, but the proposed method densely guides the whole distilled knowledge to the target network. In the end, the closer we are to student learning, the more educators we have, e.g., TAs and the teacher. Therefore, the final student model can get more opportunities to achieve better results. 

For stochastic learning of a student model, we randomly remove a fraction of the guided knowledge from trainers during the student training, which is inspired from~\cite{dropout, GaoECCV2016}. Eventually, the student network is taught from trainers ensembled slightly different for each iteration; this acts as a kind of regularization to solve the problem of overfitting which can often occur when a simple student learns from a complex teacher groups. 

The major contributions of this paper are summarized as follows: 
\begin{itemize}
    \item We propose a DGKD that densely guides each TA network with the higher-level TAs as well as the teacher and it helps to alleviate the error avalanche problem whose probability of occurrence increases as the number of TAs increases. 
    \item We revise a stochastic DGKD learning algorithm to train the student network from the teacher and multiple TAs efficiently.
    \item We demonstrate the significant accuracy improvement gained by the proposed method over well-known KD methods through extensive experiments on various datasets and network architectures. 
\end{itemize}

\section{Related Work}
\label{sec:RelatedWork}
\textbf{Knowledge Distillation:} Knowledge distillation is a popular research topic in the field of model compression \cite{Pruning, NISP2018}. We can extract distilled knowledge from a teacher network and transfer it to a student network to mimic the teacher network. The basic concept of knowledge distillation \cite{HintonKD} is to compress the knowledge of a deeper or larger model to a single computational efficient neural network. After this study, extensive research was conducted on knowledge distillation. 
Remero et al.~\cite{FitNets} introduced the transfer of a hidden activation output and Zagoruyko et al.~\cite{Zagoruyko2017} proposed transferring attention information as knowledge. Yim et al.~\cite{FSP} defined distilled knowledge from the teacher network as the flow of the solution process (FSP), which is calculated as the inner product between feature maps from two selected layers. 

Recently, Tung et al.~\cite{Tung2019} introduced similarity-preserving knowledge distillation guided training of a student network such that input pairs that produce similar activations in the teacher network produce similar activations in the student network. Zhang et al.~\cite{Linfeng2019} proposed self-distillation in which student networks train the knowledge by themselves from deeper to shallower layers so that a teacher network is not required. Because the training of the pre-trained teacher model is not required, the time for training the student model can be reduced.
Contrarily, Shen et al.~\cite{MEAL2019} believed that a student network can learn knowledge efficiently from an ensemble of teacher networks; they proposed the use of an adversarial-based learning strategy with a block-wise training loss.
For an online distillation framework, Zhang et al.~\cite{Mutual} suggested that peer students learn from each other through the cross-entropy loss between each pair of students. Chen et al.~\cite{Defang2020} also suggested using peers where multiple student models train each student model based on auxiliary peers and one group leader. Guo et al.~\cite{Quishan2020} proposed collaborative learning-based online knowledge distillation that trained students without a teacher where knowledge is transferred among arbitrary students during collaborative training. 

There have also been recent attempts to break away from the traditional method. Xu et al.~\cite{SELF2020} showed that contrastive learning as a self-supervision task helps to gain more rounded knowledge from a teacher network. Yuan et al.~\cite{Yuan2020} addressed the time-consuming training procedure of the traditional knowledge distillation methods using the few-sample where the teacher model was compressed and the student-teacher model was aligned and merged with additional layers.

\textbf{Large Capacity Gap Between Teacher and Student:} There are contrasting views regarding whether a good teacher always teaches a student well. Cho and Hariharan~\cite{JangHyun2019} found that knowledge distillation cannot succeed when the student model capacity is too low to mimic the teacher model; they presented an approach to mitigate this issue by stopping teacher training early to recover a solution more amenable for the student model. Similar to~\cite{Furlanello2018}, Jin et al.~\cite{Xiao2019} also constructed a gradually mimicking sequence of learning the teacher, which supervised the student with the teacher's optimization route. 
From a similar perspective, Mirzadeh et al.~\cite{TAKD2020} insisted that the student network performance may decrease when the capacity gap between the teacher and the student is large and introduced the multi-step knowledge distillation that employs the intermediate TA to bridge the gap between the teacher and student network. They showed that more distillation steps make a better student, up to three steps. However, considering resource constraints, they added that even one step can be effective.

Our proposed method differs from existing methods as we densely guide the student network using all assistant models generated along the way from the teacher to the student for overcoming the large gap between them. Note that our approach does not simply rely on a single model to teach the student, but uses all models that gradually become similar to the teacher's characteristics; it will be helpful to avoid the error avalanche problem.

\section{Densely Guided Knowledge Distillation using Teacher Assistants}
\label{sec:DGKD}
\subsection{Background}
The key concept of knowledge distillation~\cite{HintonKD} is training the student network to mimic the teacher network output. To achieve this goal, the fully connected layer output, logits, is used as the knowledge of the network. The loss $L_{KD}$ of the Kullback–Leibler (KL) divergence consists of the softened output of the teacher and student networks, defined as follows:
\begin{equation}
    L_{KD}=\tau^2 KL(y_S,y_T),
    \label{eq:01}
\end{equation}
where $\tau$ is the temperature parameter to control the softening of signals. $z_S$ and $z_T$ refer to the teacher and student logits, respectively and each network’s output is $y_S=softmax(z_S/\tau)$ and $y_T=softmax(z_T/\tau)$.

With the distillation loss, from equation~(\ref{eq:01}), to learn the original supervision signal, the cross-entropy loss $L_{CE}$ needs to be added with the label $y$ as follows: 
\begin{equation}
    L_{CE} = H(softmax(z_S),y).                  
    \label{eq:02}
\end{equation}

As a result, the final loss function of the conventional KD is written with the balancing parameter $\lambda$ as follows:
\begin{equation}
    L=(1-\lambda) L_{CE}+\lambda L_{KD}.                      
    \label{eq:03}
\end{equation}


\subsection{Proposed Method}

When the teacher’s performance is good, the main purpose of KD is basically a trial to guide the student knowledge at a level close to that of a teacher. However, when the gap between a student and a teacher is large in terms of the size of weight parameter or the number of layers, the best teacher does not always guide the student properly ~\cite{JangHyun2019}. To solve this problem, TAKD~\cite{TAKD2020} was proposed using intermediate-sized networks such as a TA to bridge the large gap between a teacher and a student. TAKD improved the student learning efficiency by sequentially deploying the TAs from the teacher to the student. However, TAs are smarter than the student but worse than the teacher; this eventually becomes an obstacle to further student learning due to the limited knowledge of TAs. In the end, for teaching the student well, we need a smart teacher but it is a contradiction that most of good teacher networks have many parameters, which creates a network gap with the student at the same time. 

\begin{figure}[t]
\centering
\includegraphics[
  width=7cm,
  keepaspectratio,
]{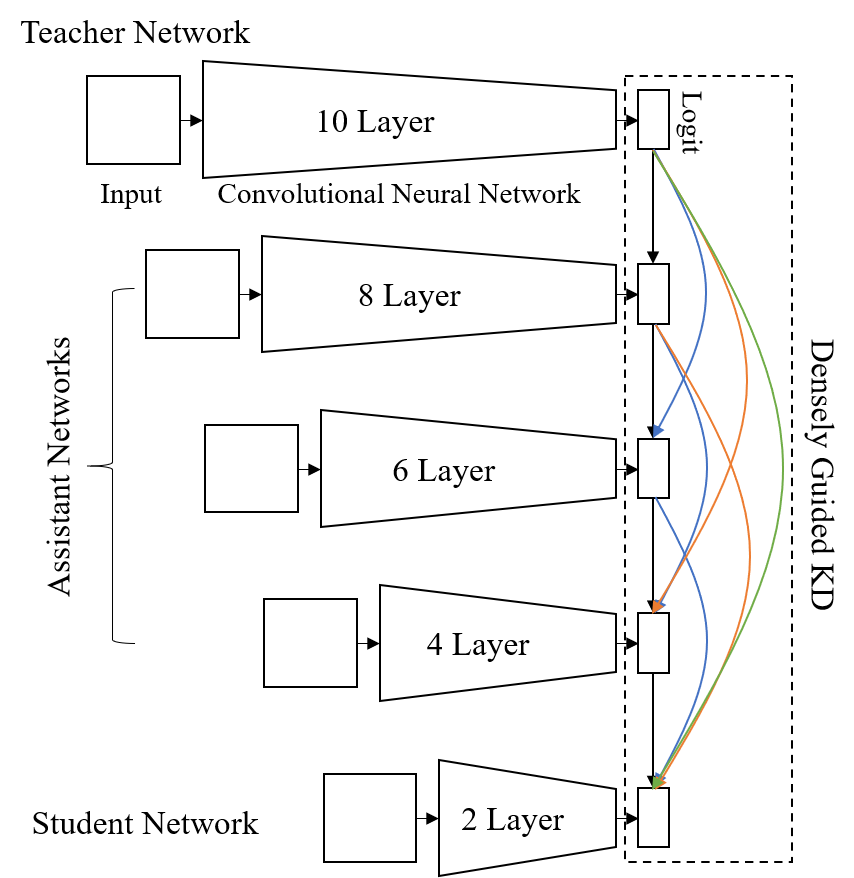}
\caption{\textbf{Overview of the proposed method.} Our densely guided knowledge distillation using multiple teacher assistant networks is able to train a small-sized student network from a large-sized teacher network efficiently through the multiple teacher assistant networks.}
\label{fig:02}
\vspace{-4mm}
\end{figure}

\textbf{Densely Guided Knowledge Distillation:} In this paper, to overcome this contradiction and achieve better performance of a shallow student network, we propose densely guided knowledge distillation using multiple TAs which are trained sequentially. As shown in Figure~\ref{fig:02}, we make use of the distilled knowledge from intermediate-sized TAs and the teacher for teaching the student network. Moreover, this densely connected distillation form is also used when teaching TAs. Note that when designing the proposed training framework for KD, the underlying idea came from DenseNet~\cite{DenseNet} which is a densely connected residual network architecture for a classification task. 

We can use several distillation losses between the assistant models and the student. For easy understanding, if there are two TA models $A_1$ and $A_2$ with the teacher model $T$, the loss of each TA can be written as follows:
\begin{equation}
\begin{split}
    L_{A_1}&=L_{T\rightarrow A_1}, \\
    L_{A_2}&=L_{T\rightarrow A_2}+L_{A_1\rightarrow A_2},\\
\end{split}
    \label{eq:04}
\end{equation}
where the right arrow at the subscript indicates the teaching direction. The student's loss $L_S$ is finally derived as follows:
\begin{equation}
    L_S=L_{T\rightarrow S}+L_{A_1\rightarrow S}+L_{A_2\rightarrow S},
    \label{eq:05}
\end{equation}
where we guide the student network using the distilled knowledge from two TAs and a teacher. Equation (\ref{eq:05}) can be expressed in the same form as equation (\ref{eq:03}) as follows:
\begin{equation}
\begin{split}
    L_S = &(1-\lambda_1 ) L_{CE_S} + \lambda_1 L_{KD_{T\rightarrow S}}+ \\
        &(1-\lambda_2 ) L_{CE_S} + \lambda_2 L_{KD_{A_1 \rightarrow S}}+\\
        &(1-\lambda_3 ) L_{CE_S} + \lambda_3 L_{KD_{A_2 \rightarrow S}}.
\end{split}
    \label{eq:06}
\end{equation}

If there are $n$ assistant models and assuming that $\lambda$ retains the same value for simplicity, the general form of the total loss is derived as the follows:
\begin{equation}
    L_S = (n+1)(1-\lambda) L_{CE_S}+
        \lambda(L_{KD_{T\rightarrow S}} +\sum_{i = 1}^{n}L_{KD_{A_{i\rightarrow S}}}).
    \label{eq:07}
\end{equation}


We distill complementary knowledge from each assistant network that has been previously learned and teach the student network with combined knowledge from all teacher assistant models. Consequently, the student network tries to mimic the various logit distributions ranging from the larger teacher network to the smaller TA network, resulting in improvement of the learning efficiency of the student network even with a large gap.

\textbf{Stochastic DGKD:} 
For efficient learning with the proposed DGKD, we adopt a stochastic learning strategy~\cite{dropout, GaoECCV2016} that randomly cuts the knowledge connections between many TAs and a student for each sample or mini-batch; we named our method as stochastic DGKD. Learning with a stochastic strategy is based on a simple intuition. We have multiple assistant models for teaching the shallow student network with a large gap, which would cause an overfitting problem due to the complex logit distribution of the TA ensemble as well as the teacher. It is necessary that the knowledge connection from the teachers is irregularly altered during the training, thus we randomly select the combination of distilled knowledge from the complex teacher and teacher assistant models. This process acts as a regularization function and it alleviates the overfitting problem. If the number of teacher and teacher assistants increases, it can relatively make the learning procedure simple.

\begin{figure}[t]
\centering
\includegraphics[
  width=7cm,
  keepaspectratio,
]{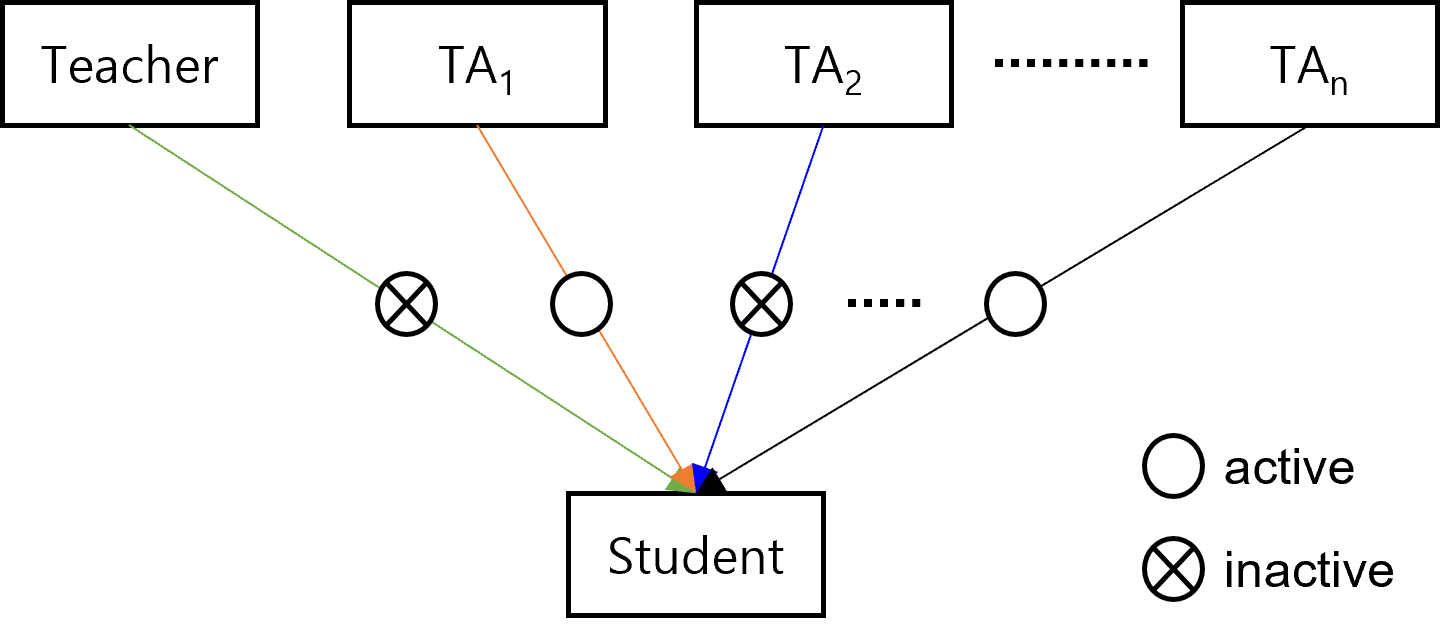}
\caption{\textbf{Concept of the stochastic DGKD method.} When having $n$ TAs and the teacher network, the student network is distilled by them. Depending on the survival probability, some of them could be randomly dropped out from a group of teachers at every training iteration.}
\label{fig:SDGKD}
\vspace{-4mm}
\end{figure}

For this purpose, we set $b_i\in\{0,1\}$ as a Bernoulli random variable and $b_i=1$ is active and $b_i=0$ is inactive during KD learning. The survival probability for Bernoulli random variable is noted by $p_i=Pr(b_i=1)$. Eventually, Equation~(\ref{eq:07}) is updated by replacing $L_{KD}$ with $b_i\cdot L_{KD_i}$.
In this paper, we use a simple dropping rule as shown in Figure~\ref{fig:SDGKD}. The possibility of dropping knowledge is equal from the teacher to the last TA and it is only applied when teaching the student because the last student has the most enough candidates to teach the distilled knowledge to the target.  
In the experiment section, we perform a more detailed empirical comparison of the different survival probabilities. 

%

\section{Experimental Setting}
\label{sec:Experiment}
\textbf{Datasets:} For fair comparisons, we evaluate the KD methods using CIFAR~\cite{CIFAR} and ImageNet~\cite{ImageNet} datasets, which are widely used as benchmarks for image classification. The CIFAR datasets comprise RGB images of size 32$\times$32 with 50,000 training images and 10,000 testing images. There are two kinds of datasets, CIFAR-10 and CIFAR-100, which have 10 and 100 classes, respectively. The ImageNet dataset contains 1,281,167 images from 1,000 classes for training and 50,000 images for validation.

\textbf{Networks:} 
We performed experiments using various networks: plain CNN and ResNet~\cite{ResNet}, WideResNet (WRN)~\cite{WideResNet}, and VGG~\cite{VGG}.
In this paper, the baseline architecture is a plain CNN, which is a VGG-like network. Following the experimental protocols of TAKD, we use a plain CNN architecture with a 10 layer-based teacher model, TA models with 8, 6, 4 layers, and a 2 layer-based student model. For a detailed comparison, we use 9, 7, 5, and 3 layer-based assistant models by removing the last convolution layer of layers 10, 8, 6, and 4. 

\textbf{Implementation Details:} We set the implementation setting with a preprocessing, optimization, training plan, and learning rate, etc., using PyTorch~\cite{Paszke2017}. We first used a random crop and random horizontal flip. We normalized for ResNet, WRN, and VGG except plain CNNs as done by TAKD. We used an initial learning rate of 0.1, stochastic gradient descent (SGD) optimizer with nesterov momentum 0.9, and weight decay 1e-4 for 160 epochs. For plain CNN, we maintained the same learning rate for all epochs, but for ResNet, WRN, and VGG, we divided the learning rate by 0.1 on epochs 80 and 120. To obtain optimal performance, we use a hyperparameter optimization toolkit\footnote{Microsoft's neural network intelligence toolkit} using the same hyperparameter and seed setting, as done by TAKD. The survival probability $p$ for the stochastic DGKD is added to balance parameter $\lambda$ and temperature parameter $\tau$. We reported the performance of the classification task for all experiments.

\section{Result and Discussion}
\subsection{Ablation Study: Comparison with TAKD }

\begin{table}[t]
\begin{center}
\caption{Accuracy comparison with all distillation steps using plain CNNs (e.g., teacher $T_{10}$, teacher assistants $A_8, A_6, A_4$, student $S_2$) on CIFAR-100. The numbers denoted by * method came from the corresponding paper.}
\label{table:01}
\begin{tabular}{c|c|c}
\hline
    Step & TAKD$^*$ & DGKD\\
\hline
\hline
Teacher (10 layer) & 56.19 & 56.15\\
Student (2 layer) & 41.09 & 41.06\\
\hline
$T_{10} \rightarrow A_{8}$	&56.75	&56.72\\
$T_{10}\rightarrow A_{8} \rightarrow A_6$	&57.53	    &\textbf{60.15}\\
$T_{10}\rightarrow A_{8} \rightarrow A_6 \rightarrow A_4$ &	52.87	&\textbf{57.63}\\
$T_{10}\rightarrow A_{8} \rightarrow A_6 \rightarrow A_4 \rightarrow S_2$ &	45.14	&\textbf{48.92}\\
\hline
\end{tabular}
\end{center}
\vspace{-4mm}
\end{table}


\begin{table}[t]
\begin{center}
\caption{Accuracy comparison with all distillation steps using ResNet (e.g., teacher $T_{26}$, teacher assistants $A_{20},A_{14}$, and student $S_8$) on CIFAR-10. The numbers denoted by * method came from the corresponding paper.}
\label{table:02}
\begin{tabular}{c|c|c}
\hline
    Step & TAKD$^*$ & DGKD\\
\hline
\hline
Teacher (26 layer) &92.48	&92.44\\
Student (8 layer) &86.61	&86.56\\
\hline
$T_{26}\rightarrow A_{20}$	&-	&92.57\\
$T_{26}\rightarrow A_{20} \rightarrow A_{14}$&	91.23&	\textbf{92.15}\\
$T_{26}\rightarrow A_{20} \rightarrow A_{14} \rightarrow S_8$&	88.01&	\textbf{89.02}\\
\hline
\end{tabular}
\end{center}
\vspace{-4mm}
\end{table}

In this section, we conduct comprehensive ablation studies to demonstrate the superiority of the proposed method by directly comparing it with TAKD. Basically, we retrain the whole network and follow the experimental protocol same to~\cite{TAKD2020}. Tables~\ref{table:01} and~\ref{table:02} show that our method DGKD achieves better results compared with TAKD in all cases. For example, as shown in Table~\ref{table:01}, the student model using the plain CNN network on CIFAR-100 shows 3.78\% better performance than TAKD for the $T_{10}\rightarrow A_{8}\rightarrow A_{6}\rightarrow A_{4}\rightarrow S_2$ path. Likewise, for ResNet on CIFAR-10, shown in Table~\ref{table:02}, over 1\% improvement is achieved for the $T_{26}\rightarrow A_{20}\rightarrow A_{14}\rightarrow S_8$ path. Moreover, we verified the steady improvements for the other steps. In particular, when there is just one TA, such as the $T_{10}\rightarrow A_8\rightarrow A_6$ path in Table~\ref{table:01} and the $T_{26}\rightarrow A_{20} \rightarrow A_{14}$ path in Table~\ref{table:02}, we confirm the good improvements by our method.
Note that TAKD bridges the large gap between a teacher and a student through TAs and it is a good choice for this case. This is largely because it can play a positive role as a bridge to transfer dark knowledge sequentially. Simultaneously, it can play a negative role in error accumulation, such as error avalanche. In the extreme case, if only one TA is present, there is no way to overcome the TA's inherent error. Even if the student can learn from the label, owing to its low capacity, it cannot get through this error by itself.

\begin{table*}[t]
 \begin{center}
 \caption{Top-1 accuracy (\%) on ImageNet. The teacher network is 34 layer-based ResNet ($T_{34}$) and the student network is 18 layer-based ResNet ($S_{18}$). We only use a single TA using 26 layer-based ResNet ($A_{26}$) for simplicity.}
 \label{table:03}
 \begin{tabular}{c|c|c||c|c|c|c|c|c|c|c|c}
 \hline
 &Teacher & Student & CC    & SP    & Online KD & KD    & AT    & CRD & SSKD & TAKD & DGKD \\ 
 & & & \cite{CC} & \cite{Tung2019} & \cite{OnlineKD} & \cite{HintonKD} & \cite{Zagoruyko2017} & \cite{CRD} & \cite{SELF2020}& \cite{TAKD2020}& \\ \hline\hline
Top-1& 73.3&	69.75&	69.96&	70.62&	70.55&	70.66&	70.7&	71.38&	71.62&  71.37&\textbf{71.73}\\ \hline
Top-5&91.42&	89.07&	89.17&	89.8&	89.59&	89.88&	90&		90.49&	90.67&  90.27&\textbf{90.82}\\ \hline

 \end{tabular}
 \end{center}
 \vspace{-4mm}
\end{table*}


We conducted another comparison experiment using the large-scale dataset, ImageNet~\cite{ImageNet}, presented in Table~\ref{table:03}. In this experiment, we used a 34 layer-based teacher, a 18 layer-based student, and a single TA with 26 layers using the ResNet. Our method achieves over 1\% better accuracy than Hinton's KD and shows the best performance among CC~\cite{CC}, SP~\cite{Tung2019}, online KD~\cite{OnlineKD}, KD~\cite{HintonKD}, AT~\cite{Zagoruyko2017}, CRD~\cite{CRD}, SSKD~\cite{SELF2020}, and TAKD~\cite{TAKD2020} when the 34 layer teacher teaches the 18 layer student. As a result, we can conclude that our method works efficiently regardless of the database.

\subsection{Ablation Study: Classifier Ensemble}
\begin{table}[t]
\begin{center}
\caption{Comparison results with the ensemble methods using plain CNNs on CIFAR-100.}
\label{table:ensemble}
\begin{tabular}{c|c}
\hline
Method & Accuracy\\ 
\hline
\hline
Teacher (10 layer) & 56.15\\
Student (2 layer)& 41.06\\
\hline
KD~\cite{HintonKD}& 42.56\\
TAKD using three TAs~\cite{TAKD2020} & 45.14\\
Ensemble using four $T_{10}$s & 42.57\\
Ensemble using $T_{10}$, $T_{8}$, $T_{6}$, and $T_{4}$ & 43.25\\
DGKD & \textbf{48.92}\\
\hline
\end{tabular}
\end{center}
\vspace{-4mm}
\end{table}

We make experimental results regarding the ensemble classifiers in Table~\ref{table:ensemble}. The first ensemble is made by four different 10 layer-based teachers from scratch independently and the second ensemble is built by 10 layer-based teacher $T_{10}$, 8 layer-based teacher $T_8$, 6 layer-based teacher $T_6$, and 4 layer-based teacher $T_4$. Note that these teachers are trained independently. Both these ensembles achieved better results than the KD~\cite{HintonKD} but they fail to overcome the large gap between the teacher and student models. On the other hand, TA-based methods including ours solve this problem successfully. From this result, we can infer that the simple ensemble method is not a good solution for the large gap between the teacher and student models in a KD task. 


\subsection{Error Avalanche Problem of TAKD}
An example of an error avalanche problem is briefly explained in Figure~\ref{fig:01} (c). In the case of TAKD, the student can learn from only an upper TA independently; the TA also learns from an upper TA sequentially following the distillation path. Thus, if an upper TA model transfers incorrect knowledge to the next model, this incorrect knowledge continuously accumulates following the sequential distillation path of TAKD. 

\begin{figure}[t]
\centering
\includegraphics[width=7.5cm]{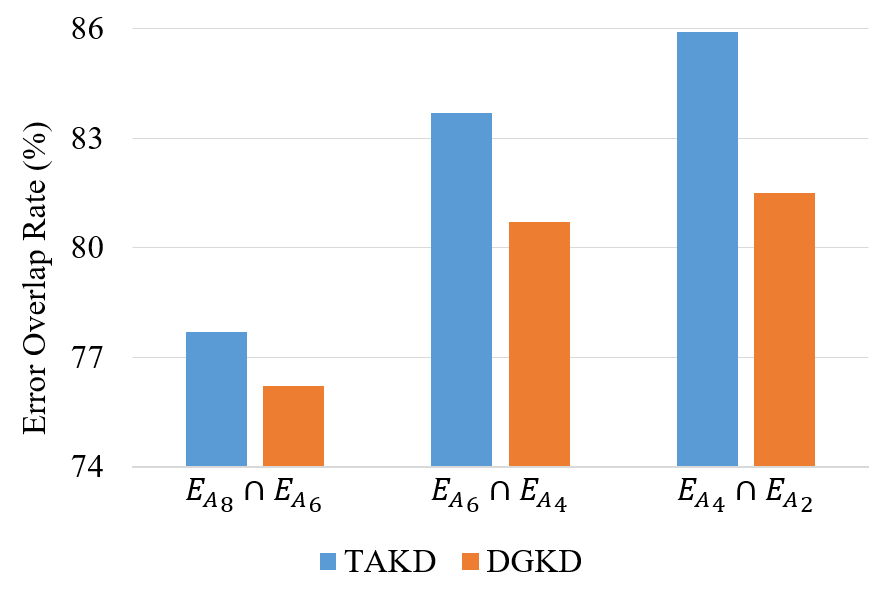}
\caption{\textbf{Error avalanche problem.} The error overlap rate indicates the intersection proportion of a higher-level model’s incorrect answer and a lower level model’s incorrect answer when we have the teacher $T_{10}$, the student $S_2$, and three TAs (e.g., $A_8$, $A_6$, and $A_4$). $E_i \cap E_j$ where $E_i$ is the $i$th plain CNN model's error examples on CIFAR-100.}
\label{fig:03}
\vspace{-4mm}
\end{figure}

We conducted an experiment to check the error overlap rate between two neighboring models using the full distillation path (e.g., $T_{10}\rightarrow A_8 \rightarrow A_6 \rightarrow A_4 \rightarrow S_2$), as shown in Figure~\ref{fig:03}, and observe that the error overlap rate of TAKD is much higher than that of DGKD in all cases. In particular, we can see that the closer to the student, the larger the gap in error overlap rates between TAKD and DGKD. From this viewpoint, we conclude that the teacher and TAs can help the student avoid the error avalanche problem by the proposed method.

\begin{figure}[t]
\centering
\includegraphics[width=8.3cm]{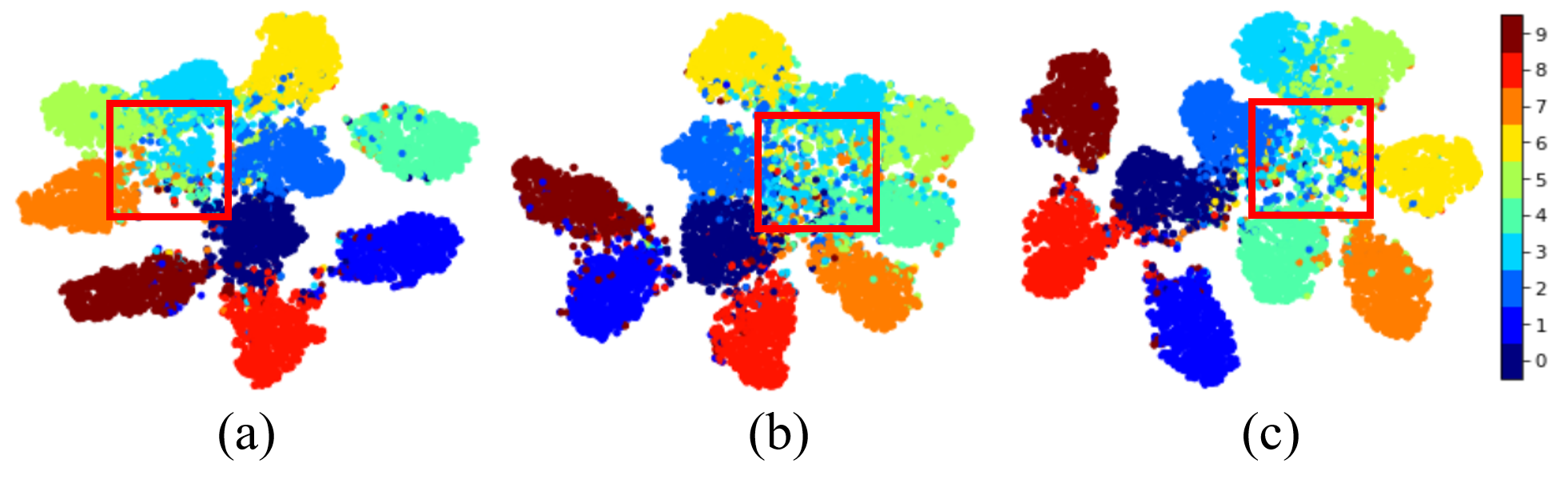}
\caption{\textbf{t-SNE visualizations} of (a) KD for $T_{26}\rightarrow A_{20}$, (b) TAKD for $A_{20}\rightarrow S_{14}$, and (c) our DGKD for $A_{20}\rightarrow S_{14}$ using ResNet on CIFAR-10. Looking at the class distribution in the red box, we can see the different error accumulation rates of (b) TAKD and (c) our DGKD.}
\label{fig:tsne}
\vspace{-2mm}
\end{figure}

We also verify this problem by t-SNE visualization at Figure~\ref{fig:tsne}, where it could be seen that the error accumulation increased more when going from TA to student networks than when going from teacher to TA networks. In this case, the student model deployed at the bottom would suffer from the accumulated errors due to the error avalanche problem. The student has a chance to learn from the cross-entropy supervision loss, but it is not very helpful to resolve the error avalanche problem because the 2 layer-based student’s learning capacity for the supervision signal is not enough to overcome this problem. Therefore, TAs sequentially deployed based on TAKD could be an insufficient solution when there is a large gap between the teacher and the student. Due to the fundamental limitations, it is difficult to expect the best performance improvement at KD. However, our DGKD teaches the student from the teacher to the TAs at the same time and this error avalanche problem could be alleviated properly.

\begin{figure}[t]
\centering
\includegraphics[width=8cm]{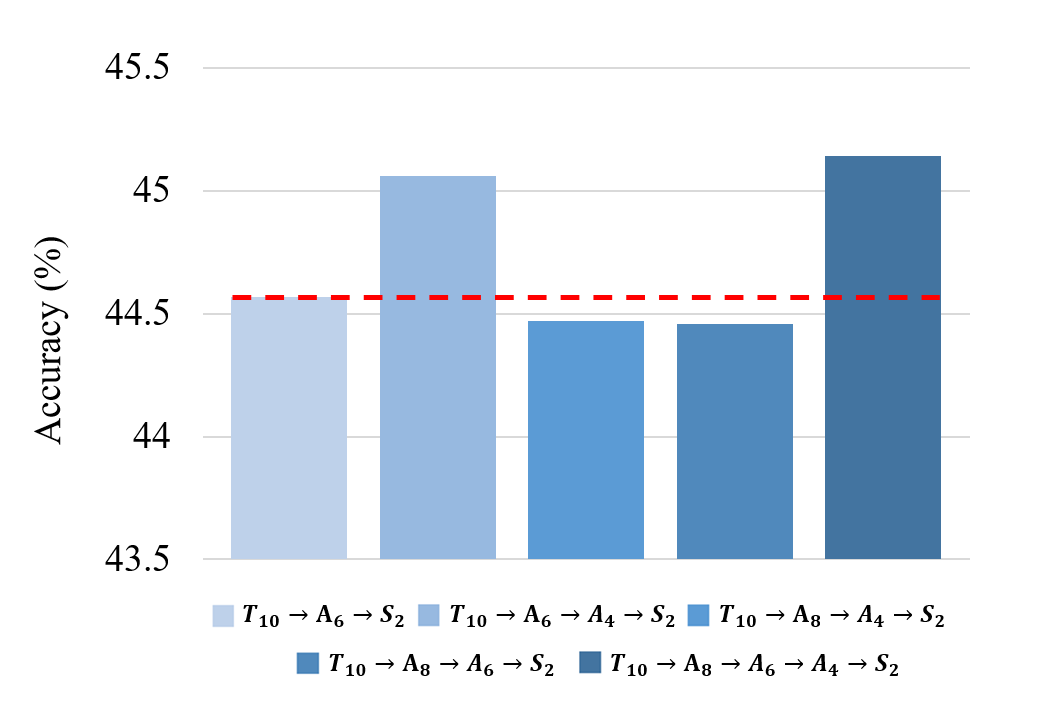}
\caption{\textbf{Various TAKD distillation paths and the corresponding results.} Different distillation paths result in different accuracies at CIFAR-100, but the deeper path based on many TAs does not always guarantee the best performance owing to the error avalanche problem.}
 \label{fig:04}
 \vspace{-4mm}
\end{figure}

\begin{table}[t]
\begin{center}
\caption{Extensive distillation path with plain CNNs on CIFAR-100 dataset by adding TAs intermediately; $n$ is the number of TAs used. 
}
\label{table:04}
\begin{tabular}{c|c|c|c}
\hline
Step & $n$	& TAKD	&DGKD\\
\hline
\hline
Teacher (10 layer)&-& \multicolumn{2}{c}{56.19}\\
Student (2 layer)&-&	\multicolumn{2}{c}{41.09}\\
\hline
$T_{10}\rightarrow A_6\rightarrow S_2$&1	&44.57&	\textbf{45.85}\\
$T_{10}\rightarrow A_8\rightarrow A_6\rightarrow A_4\rightarrow S_2$ &3&	45.14&	\textbf{48.92}\\
$T_{10}\rightarrow A_9\rightarrow A_8\rightarrow A_7\rightarrow A_6$    &\multirow{2}{*}{7}  &\multirow{2}{*}{44.07}&	\multirow{2}{*}{\textbf{49.56}}\\
$\rightarrow A_5\rightarrow A_4\rightarrow A_3 \rightarrow S_2$         &   &       & \\
\hline
\end{tabular}
\end{center}
\vspace{-6mm} 
\end{table}

\label{sec:SDGKD}
\begin{figure}[t]
\centering
\includegraphics[width=6.7cm]{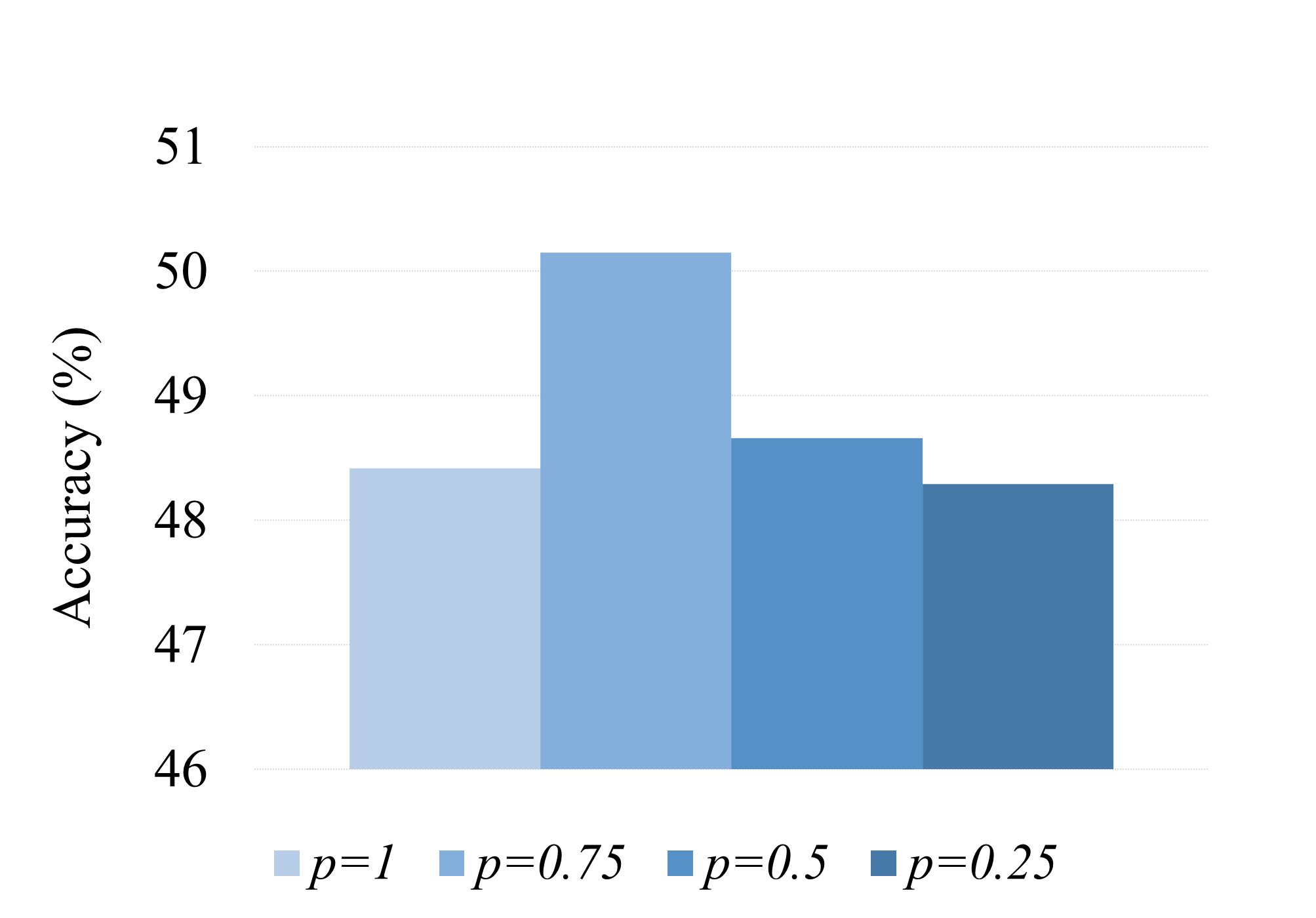}
\caption{\textbf{Stochastic DGKD.} Performances of plain CNNs with $T_{10}\rightarrow A_{8}\rightarrow A_{6}\rightarrow A_{4}\rightarrow S_{2}$ by the different survival probability $p$ at CIFAR-100.}
\label{fig:05}
\end{figure}

\subsection{Knowledge Distillation Path}
We also investigated the assertion made by~\cite{TAKD2020} that a full distillation path is optimal for KD. As shown in Figure~\ref{fig:04}, the deepest model (e.g., $T_{10}\rightarrow A_8\rightarrow A_6\rightarrow A_4\rightarrow S_2$) path is the best as described by ~\cite{TAKD2020}; however, the models with two TAs (e.g., $T_{10}\rightarrow A_8\rightarrow A_4\rightarrow S_2$ and $T_{10}\rightarrow A_8\rightarrow A_6\rightarrow S_2$) show lower performances than the model with one TA (e.g., $T_{10}\rightarrow A_6 \rightarrow S_2$) indicated by the red dash line. In this respect, we can infer that TAKD does not always improve the performance even if there are multiple TAs.

We attempted to determine whether TAKD can achieve the best accuracy even when the distillation path is extended to the maximum.
As shown in Table~\ref{table:04}, when all TAs are added intermediately for extending the distillation path to the maximum possible levels, the full distillation path model (e.g., $n=7$/44.07\%) of TAKD shows an even lower performance than the other models (e.g., $n=3$/45.14\% and $n=1$/44.57\%). In this respect, we could be sure that the error avalanche problem would occur in this case. As a result, when the path is deeper and deeper, the error avalanche problem can get worse than expected. 
However, using the proposed DGKD, the accuracy gradually improves when the distillation path becomes deeper, ranging from $n=1$ to $n=7$; a full distillation path (e.g., $n=7$) achieves the best accuracy of 49.56\%. Note that as the path lengthens, the performance of our DGKD method improves unlike TAKD, and it was over approximately 5\% better in accuracy compared with TAKD.

In summary, by adapting our proposed DGKD method which uses all higher-path model teacher and TAs together, a low capacity student can overcome the error avalanche problem with the proper trainers.

\begin{table}[t]
\caption{Stochastic learning-based DGKD ($p=0.75$) comparison results using a 2 layer-based plain CNN student with path $T_{10}\rightarrow A_{8}\rightarrow A_{6}\rightarrow A_{4}\rightarrow S_{2}$ on CIFAR-100, and an 8 layer-based ResNet student with path $T_{26}\rightarrow A_{20}\rightarrow A_{14}\rightarrow S_8$ on CIFAR-10.}
\label{table:05}
\begin{center}
{\small
\begin{tabular}{c|c|c|c|c}
\hline
Model &Dataset&	TAKD&	\multicolumn{2}{c}{Ours}\\
\cline{4-5}
&&& DGKD & Stochastic \\
&&&&DGKD \\
\hline
\hline
PlainCNN& CIFAR-100&	45.14&	48.92&	\textbf{50.15}\\
ResNet& CIFAR-10&	88.01&	89.02&	\textbf{89.66}\\
\hline
\end{tabular}
}
\end{center}
\vspace{-4mm}
\end{table}

\subsection{Stochastic DGKD}
We proposed the stochastic learning-based DGKD for further performance improvements of a student. Specifically, our stochastic DGKD is directly inspired by the dropout concept. If there are $n$ TAs, the student can learn from the $n+1$ knowledge corpus including a teacher. We randomly removed the connections between the student and the trainers (the teacher and TAs) based on the survival probability $p$.
To figure out the tendency of the survival probability, we performed the experiments shown in Figure~\ref{fig:05}. When the survival probability is $p=0.75$, the student performance shows the best accuracy.
Note that the stochastic DGKD with the survival probability $p$ from 0.75 to 0.5 shows the better accuracy than the vanilla DGKD $p=1$ but at $p=0.25$, the accuracy is slightly below than the vanilla DGKD because we only have four knowledge connections in this experiment and dropping three results in losing the chance to learn the proper knowledge from the number of trainers. 
In this respect, we conclude that the survival probability $p=0.75$ is the optimal hyper-parameter when we have three TAs and a teacher for the stochastic DGKD. 

\begin{table*}[t]
\begin{center}
\caption{Comparison with well-known KD methods using ResNet on CIFAR-10. We used 26 layer-based ResNet as the teacher model which teaches two different student models, e.g., 8 layer-based ResNet and 14 layer-based ResNet, respectively. For TAKD and our DGKD, we used knowledge distillation paths, $T_{26}\rightarrow A_{20}\rightarrow A_{14}\rightarrow S_{8}$ and $T_{26}\rightarrow A_{20}\rightarrow S_{14}$, respectively.}
\label{table:06}
{\small
\begin{tabular}{c|c|c|c|c|c|c|c|c|c}
\hline
 & Student & KD~\cite{HintonKD} & FitNet~\cite{FitNets} & AT~\cite{Zagoruyko2017} & FSP~\cite{FSP} & BSS~\cite{BSS} & Mutual~\cite{Mutual} & TAKD~\cite{TAKD2020} & Ours\\
\hline
\hline
ResNet8	&86.02&	86.66&	86.73&	86.86&	87.07&	87.32&	87.71&	88.01&	\textbf{89.66}\\
ResNet14 &	89.11&	89.75&	89.82	&89.84&	89.92&	90.34&	90.54&	91.23&	\textbf{92.34}\\
\hline
\end{tabular}
}
\end{center}
\vspace{-4mm}
\end{table*}


Table~\ref{table:05} shows a performance comparison among TAKD, DGKD, and stochastic DGKD. As expected, the stochastic DGKD shows the best performance on the other networks and datasets.
Specifically, the proposed stochastic DGKD using the 2 layer-based plain CNN student network achieves 1.23\% better result than the original DGKD and 5.01\% better result than TAKD on the CIFAR-100. This improvement is the same on 8 layer-based ResNet student on CIFAR-10. From this, we can conclude that the proposed method works successfully in the case of a large gap between the teacher and student network.

\begin{table}[t]
\begin{center}
\caption{Comparison with the previously published KD methods using WRN, ResNet, and VGG on CIFAR-100. A bold number is a best accuracy, and an underlined number is a second best one. }
\label{table:07}
\begin{tabular}{c|c|c|c}
\hline
Teacher & WRN40$\times$2 & ResNet56    &VGG13\\
Student & WRN16$\times$2 & ResNet20    &VGG8\\
\hline
\hline
Teacher & 76.46 &73.44  &75.38\\
Student &73.64  &69.63  &70.68\\
\hline
KD~\cite{HintonKD}      &74.92  &70.66  &72.98\\
FitNet~\cite{FitNets}  &75.75  &71.60  &73.54\\
AT~\cite{Zagoruyko2017}&  75.28&  \underline{71.78}   &73.62\\
SP~\cite{Tung2019}&  75.34  &71.48  &73.44\\
VID~\cite{VID}&74.79&71.71&73.96    \\
RKD~\cite{RKD}&75.40&71.48&73.72    \\
PKT~\cite{PKT}&76.01&71.44&73.37    \\
AB~\cite{AB}& 68.89&71.49&74.27\\
FT~\cite{FT}& 75.15&71.52&73.42\\
CRD~\cite{CRD}&\underline{76.04}&71.68&74.06    \\
SSKD~\cite{SELF2020}&\underline{76.04}&71.49&\textbf{75.33}\\
TAKD~\cite{TAKD2020}&   75.04&70.77&73.67\\
\hline
Ours& \textbf{76.24}&\textbf{71.92}&\underline{74.40}\\
\hline
\end{tabular}
\end{center}
\vspace{-4mm}
\end{table}

\subsection{Comparison with The State-of-the-art Methods}

To verify the generality of the proposed method, we compared its performance with the well-known KD methods~\cite{HintonKD, FitNets, Zagoruyko2017, FSP, BSS, Mutual, TAKD2020}.
As summarized in Table~\ref{table:06}, the proposed DGKD achieves the best performances compared with the well-known KD methods such as KD~\cite{HintonKD}, FitNet~\cite{FitNets}, AT~\cite{AB}, FSP~\cite{FSP}, BSS~\cite{BSS}, Mutual~\cite{Mutual}, and TAKD~\cite{TAKD2020} at both ResNet-based student models with 8 and 14 layers, respectively at CIFAR-10.
For comparison results with various backbone architectures, for example, WRN~\cite{WideResNet}, ResNet~\cite{ResNet}, and VGG~\cite{VGG}, Table~\ref{table:07} shows that our proposed method performed favorably against the state-of-the-art KD methods.
Specifically, a 40$\times$2 layer-based WRN, a 56 layer-based ResNet, and a 13 layer-based VGG as teachers, and the corresponding students are a 16$\times$2 layer-based WRN, a 20 layer-based ResNet, and an 8 layer-based VGG, respectively. In this experiment, we also use different numbers of the teacher assistant models for the different networks. In detail, we used the following knowledge distillation paths: 
$T_{40\times 2}\rightarrow A_{34\times 2}\rightarrow A_{28\times 2}\rightarrow A_{22\times 2}\rightarrow S_{16\times 2}$ for WRN, $T_{56}\rightarrow A_{44}\rightarrow A_{32}\rightarrow S_{20}$ for ResNet, and $T_{13}\rightarrow A_{11}\rightarrow S_{8}$ for VGG. As shown in Table~\ref{table:07}, when the difference between the teacher and student networks is large (e.g., from WRN40-2 to WRN16-2 and from ResNet56 to ResNet20) our method shows the best accuracy among the state-of-the-art methods, but when the difference is not large (e.g., from VGG13 to VGG8), our method shows the second-best accuracy.

From the results of all these experiments, we can conclude that the proposed method shows the best performance not only when the gap between the teacher and the student is large but also compared with the general KD methods.

\section{Conclusion}
\label{sec:Conclusion}
In this paper, we proposed a densely guided knowledge distillation using multiple assistants to improve the performance of the student with low capacity compared to the teacher. Empirically, we found that the error avalanche problem can easily occur as the assistant knowledge distillation paths deepen. When there are multiple assistants, if the upper assistant transfers the wrong answers to the next assistant and it continues recursively, it can be difficult for the student to avoid error avalanche problems because of its low capacity. Thus, we proposed the novel method using the knowledge of the teacher and the whole assistants to provide more opportunities for the student to learn the right answers during the training. Our experiments demonstrate that our proposed method can play a key role in resolving the error avalanche problem. Moreover, for efficient student learning, we adapted the stochastic learning role by randomly abandoning the teacher or assistant knowledge. Using this strategy, our proposed method achieves the state-of-the-art among the well-known distillation methods. We believe that our proposed method can boost planning the distillation path deeper and deeper using the multiple TAs, which improving the performances of the low capacity-based student networks in the real world applications.


\end{document}